\documentclass[conference]{IEEEtran}
\IEEEoverridecommandlockouts
\usepackage{cite}
\usepackage{amsmath,amssymb,amsfonts}
\usepackage{graphicx}
\usepackage{subfigure}
\usepackage{adjustbox}
\usepackage{algpseudocode}
\usepackage{algorithm}
\usepackage{multicol}
\usepackage{booktabs}
\usepackage{multirow}
\usepackage[english]{babel}
\usepackage{xcolor}
\usepackage{textcomp}
\usepackage{xcolor}
\usepackage{soul}
\def\BibTeX{{\rm B\kern-.05em{\sc i\kern-.025em b}\kern-.08em
    T\kern-.1667em\lower.7ex\hbox{E}\kern-.125emX}}
\begin{document}

\title{MicroXercise:  A Micro-Level Comparative and Explainable System for Remote Physical Therapy}

\author{%
\IEEEauthorblockN{Hanchen David Wang$^{*}$, Nibraas Khan$^{*}$, Anna Chen$^{*}$, Nilanjan Sarkar$^{\dagger}$, Pamela Wisniewski$^{*}$, Meiyi Ma$^{*}$}
\IEEEauthorblockA{
{$^{*}$Computer Science}, {$^{\dagger}$Mechanical Engineering} \\
Vanderbilt University, Nashville, TN \\
\{hanchen.wang.1, nibraas.a.khan, anna.chen, nilanjan.sarkar, pam.wisniewski, meiyi.ma\}@vanderbilt.edu \\
}
}


\maketitle

\newcommand{\revise}[1]{\textcolor{blue}{#1}}
\renewcommand{\revise}[1]{{#1}}

\begin{abstract}
Recent global estimates suggest that as many as 2.41 billion individuals have health conditions that would benefit from rehabilitation services. Home-based Physical Therapy (PT), faces significant challenges in providing interactive feedback and meaningful observation for therapists and patients. To fill this gap, we present MicroXercise, which integrates micro-motion analysis with wearable sensors, providing therapists and patients with a comprehensive feedback interface, including video, text, and scores. Crucially, it employs \revise{multi-dimensional Dynamic Time Warping (DTW) and} attribution-based explainable \revise{methods to analyze the existing deep learning neural networks in monitoring exercises}, focusing on a high granularity of exercise. This \revise{synergistic} approach is pivotal, providing output matching the input size to precisely highlight critical subtleties and movements in PT, thus transforming complex AI analysis into clear, actionable feedback. By highlighting these micro-motions in different metrics, such as stability and range of motion, MicroXercise significantly enhances the understanding and relevance of feedback for end-users. Comparative performance metrics underscore its effectiveness over traditional methods, such as a 39\% and 42\% improvement in Feature Mutual Information (FMI) and Continuity. MicroXercise is a step ahead in home-based physical therapy, providing a technologically advanced and intuitively helpful solution to enhance patient care and outcomes.

\end{abstract}


%

\section{Introduction}

Estimates suggest that, globally, as many as 2.41 billion individuals have health conditions that would benefit from rehabilitation services, however, a person’s access to rehabilitation, and their ability to adhere to treatment may be hindered by many factors, including costs, travel needs and lost worktime to attend multiple on-site visits, and the patient’s inability to perform their program independently. If a patient is not independent in their home exercise program, they may complete exercises incorrectly between on-site visits, potentially leading to unnecessary pain or slow recovery. Wearable technology that can monitor the patient’s home program performance and provide real-time feedback on exercise quality would facilitate adherence and improve treatment outcomes. Advances in technology are urgently needed to develop innovative, accessible, and sustainable techniques that facilitate a person’s participation in rehabilitation. 

Existing off-site monitoring systems can only detect the types of exercise performed or calculate calories burned, which are insufficient for capturing the quality of the exercise or assessing the patients' progress. Moreover, modern deep learning-based solutions for healthcare often do not explain their results sufficiently, making it difficult for patients and therapists to understand and trust the results. 
Artificial Intelligence (AI) techniques have seamlessly integrated into our daily lives, influencing many decisions, from mundane choices to critical healthcare recommendations. In particular, the healthcare domain has seen a growing influence of AI, with systems being developed for recognizing various diseases such as skin, breast, and brain tumors~\cite{nadeem_brain_2020, houssein_deep_2021, dreiseitl_comparison_2001}. Its application in healthcare is particularly significant, given the increasing need for precise and efficient treatments. 
This is especially true in the domain of Physical Therapy (PT), where an intricate understanding of human motion is crucial \cite{burns_shoulder_2018, osullivan_telehealth_2022}. PT interventions—including passive restorative, exercise, and advice—aim to improve mobility, alleviate pain, and ultimately enhance patient outcomes. Traditional methods of evaluating and treating patients in PT often rely on the clinician's experience and observational skills, which could be subjective and lack personalization for large-scale applications. The COVID-19 pandemic has further intensified the importance of PT, particularly for at-home exercises and rehabilitative care. Reflecting this, the PT domain is set to witness substantial growth, with projections suggesting a 17\% increase in employment over the next decade \cite{tmcconnell_2020_2020, us_bureau_of_labor_statistics_physical_2022}. This rising demand further underscores the need for personalized AI solutions in remote PT.

In light of the expanding PT workforce, human activity recognition (HAR) has emerged as a key AI component for personalized treatment \cite{morris_recofit_2014}. Specifically, utilizing lightweight sensor technology such as an inertial measurement unit (IMU), HAR is capable of identifying and categorizing various human motions and activities \cite{zhang_wearsign_2022, santhalingam_synthetic_2023}. In a PT context, the sensory data generated by IMU can monitor a patient's movements over time to provide quantitative, accurate traits. This can then inform the treatment plan, offering a more personalized approach to care and potentially leading to more effective and faster rehabilitation of patients.

However, the adoption of AI-based HAR in clinical settings faces challenges, primarily due to the ``black box'' nature of many AI algorithms. While these models are effective in making predictions or classifications, they often lack transparency in how they arrive at these conclusions. This opacity can be a significant drawback in healthcare applications, where understanding the reasoning behind a diagnosis or treatment recommendation is crucial \cite{rojat_explainable_2021, singh_explainable_2020}. This is where {eXplainable AI (XAI)} comes into the picture. XAI aims to make the decision-making processes of AI models transparent and understandable, allowing clinicians and doctors to trust the technology and better integrate it into their practice.

Various techniques have been developed to achieve this level of transparency, primarily in the realm of attribution-based methods. Notable among these are Gradient-weighted Class Activation Mapping (Grad-CAM) \cite{selvaraju_grad-cam_2020}, Saliency Maps \cite{simonyan_deep_2014}, and Integrated Gradients (IG) \cite{sundararajan_axiomatic_2017}. These methods highlight the significance of individual input features in determining the model's final output. In essence, they offer a mapping that quantifies the contribution of each input attribute to the decision-making process. This category of techniques, often referred to as `heatmap visualization' \cite{tjoa_survey_2021}, allows for more transparent interpretations by revealing what features the model deems crucial in its computations.


\subsection{Motivation}

Two limitations in existing systems impact their utility in physical therapy applications, \revise{particularly, for shoulder PT}. First, current XAI methods, such as heatmap visualization techniques, are often non-intuitive and not explainable for end-users, limiting their usefulness in real-world scenarios \cite{lopes_xai_2022}. Second, these methods commonly provide a holistic view of activity but fall short in isolating specific moments (or micro-motions) that require user modification---such as raising an arm higher during a shoulder abduction exercise in the first half of the exercise or asking users to slow down in the lowering arm part of the exercise \cite{mosqueira-rey_human---loop_2023}. These limitations necessitate more user-friendly and targeted XAI-heatmap approaches for enhancing the applicability of HAR models in \revise{shoulder PT} \cite{slijepcevic_explaining_2022}.

\subsection{Challenges}

The first substantial challenge lies in translating attribution-based methods into actionable insights for end-users. While these techniques, such as Saliency and IG, offer a way to ``interview'' the deep \revise{neural network} model to understand its decision-making processes, they often don't translate easily into practical, real-world advice. A potential technique is the comparison of a current workout (referred to as the ``signal exercise'') and an example or ideal exercise (referred to as the ``anchor exercise''), as presented by previous research utilizing \revise{spatiotemporal} Siamese Neural Networks (SNN) \cite{wang_physiq_2023}. However, even with such sophisticated models, given the heatmap visualization, the challenge remains: how do we translate these high-level comparisons into clear, actionable feedback for the end-user?

The second challenge is related to the intricate task of isolating and understanding micro-motions within physical activities. While breaking down exercises into smaller, more manageable components could be beneficial, doing so in a meaningful way that preserves the context of the overall exercise is not trivial. For example, directing a user to raise their arm higher during only the first half of a shoulder abduction exercise would require an acute focus on that specific micro-motions without losing sight of the larger activity.

\subsection{Contributions}
    We create MicroXercise, an innovative system that leverages a micro-motion algorithm that generates comprehensive and presentable feedback, demonstrated in multiple modalities, including text, avatars, and video highlights, utilizing the capabilities of the \revise{existing neural network, particularly, on spatiotemporal SNN,} with attribution-based methods for remote \revise{shoulder} PT. This approach enriches the user's understanding and engagement, overcoming the non-intuitive nature of existing heatmap visualization techniques.
    
     \revise{Our additional technical contribution is the development of a new multi-dimensional Dynamic Time Warping (DTW) model that works in synergy with existing attribution-based methods, specifically tailored to segment and analyze single-repetition exercise comparison results through existing neural networks.} This fusion allows for the extraction of granular, micro-motion-level insights, thereby fine-tuning the feedback and making it more actionable for end-users in the physical therapy domain.
    
    Our experiment conducted a detailed evaluation of explanation methods using three metrics: monotonicity, feature mutual information (FMI), and continuity. We compared three attribution methods, including a modified version against an unmodified baseline. The results revealed significant improvements with the modified method. On average, FMI improved by approximately 39\% with our modified approach. Regarding continuity, we observed an average decrease (indicating improvement) of about 42\%. These findings demonstrate our experiment's enhanced interpretability and fidelity of the modified attribution methods.

\begin{figure}[t]
    \centering
    {\includegraphics[width=0.95\columnwidth]{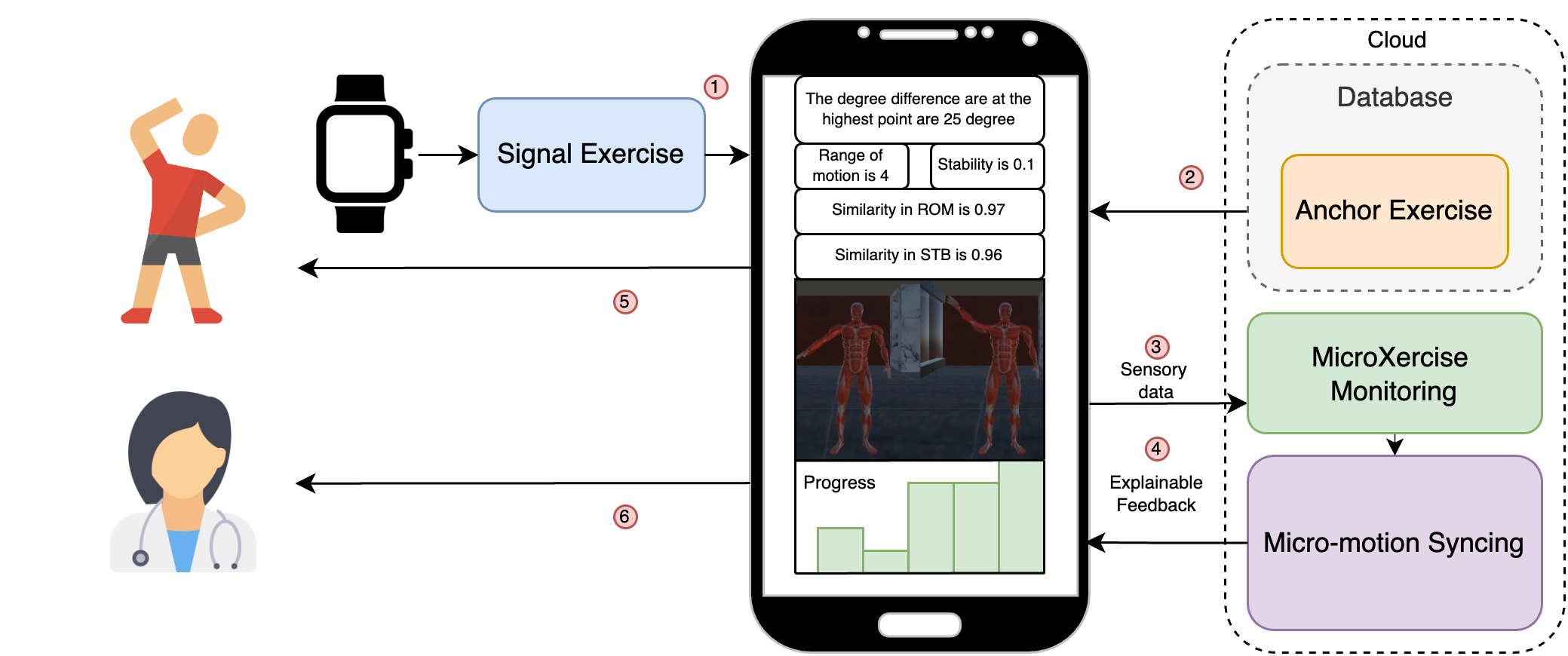}}
    \caption{This diagram illustrates the process of exercise performance and feedback generation using our system. The user performs an exercise while wearing a smartwatch, which generates a off-site exercise (or ``signal exercise") sent to the user's smartphone in (1). The smartphone accesses the off-site exercise and the on-site exercise (``anchor exercise"), which is supervised by the therapist in clinics and stored in the cloud, then processes these two exercises together and sends them to the micro-motion algorithm for analyzing in cloud (2). This provides explainable feedback, such as the range of motion and stability, and micro-motion feedback in text and visualization (3). Next, the user will have access to the generated feedback, as shown in (4). Additionally, the therapists can view users' feedback and exercise history, as shown in (5). }
    \label{fig:user_perspective}
\end{figure}

\section{System Overview of MicroXercise}\label{sec:designs}
Our system aims to elevate adherence to Human Behavior Physical Therapy (HBPT) by specifically targeting user self-efficacy and enabling patient-driven care \cite{rowland_what_2020}. Self-efficacy is fundamentally an individual's confidence in their ability to successfully carry out actions that yield desired outcomes. In the context of HBPT, it serves as a pivotal factor influencing a patient's adherence to exercise routines and overall treatment. Our system addresses the prevalent challenge where users feel unsure in utilizing mobile health (mHealth) solutions. Leveraging advanced algorithms and micro-motion analysis, we transform complex sensor data from a deep learning model into easy-to-understand and actionable feedback. \revise{Designed specifically for shoulder PT patients who are working to maintain exercise routines developed in close collaboration with their therapists, our application seeks to support long-term practices rather than serving as a replacement for acute care or professional therapy in situations with higher risks.}


Our system adopts a three-layered architecture consisting of smartwatches, smartphones, and cloud computing. Smartwatches are responsible for real-time data collection, smartphones serve as the user interface providing interactive feedback, and cloud-based systems handle computational processing and algorithmic tasks. Within this system, we have developed three main modules: MicroXercise Monitoring (Sec. \ref{subsec:xai_monitor}), Micro-motion Syncing (Sec. \ref{subsec:micro-motion}), and Generation of Explainable Feedback (Sec. \ref{subsec:gen_xai_mmm_res}). These modules are intricately designed to meet our system's objectives: delivering actionable performance guidance and enhancing user self-efficacy, as visually represented in Figure \ref{fig:user_perspective}.

In MicroXercise, this module leverages attribution-based methods, such as IG, Saliency, or DeepLIFT, to interpret our multitask SNN. 
It analyzes both the signal and anchor exercises to produce actionable and transparent outcomes. Next, acting as the core algorithmic component, Micro-motion Syncing employs signal processing and spatiotemporal DTW to scrutinize exercises at the micro-level. 
It compares user-generated signals with pre-established anchor exercises, using this data for subsequent exercise segmentation. Lastly, we transform the analytical results into both visual and textual feedback, making extensive use of avatar-generation methods. 

Visual and textual feedback are core components of the MicroXercise app, designed to offer users real-time insights into their performance and overall well-being. Visual feedback, demonstrated in Figure \ref{fig:user_perspective}, utilizes real-time graphs and charts to display performance metrics. These graphical elements enable users to track and regulate exercise intensity relative to their anchor results. Complementing the visuals, textual feedback provides granular details, such as similarity scores to anchor exercises and range of motion variations in text. This information is synthesized to prevent information overload, translating complex IMU sensor data into actionable insights, as illustrated in the generated texts on the smartphone.

\section{Explainable Exercise Performance Quantification}\label{sec:quantification}

\begin{figure}[htbp]
  \centering
  \includegraphics[width=\columnwidth]{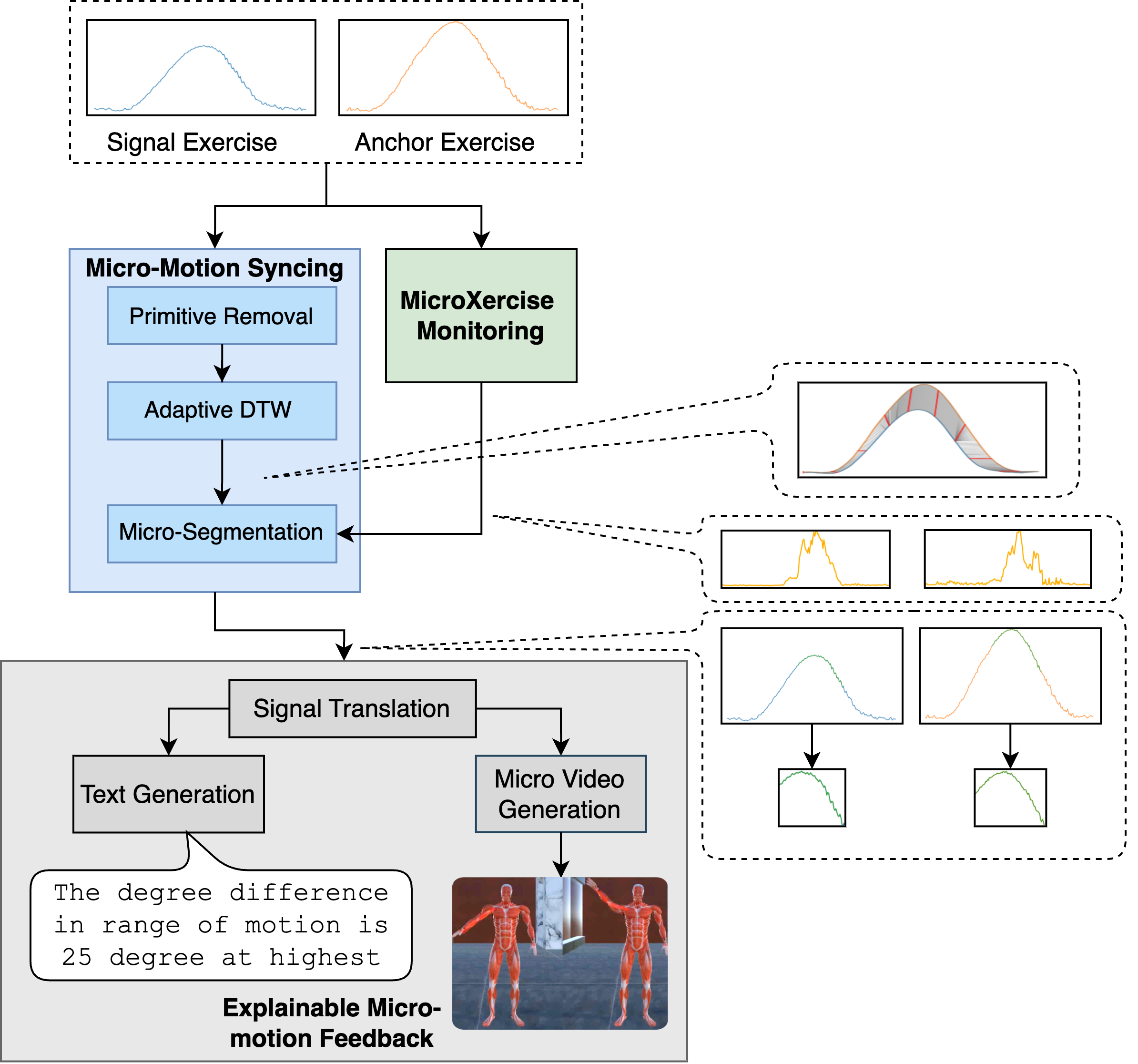}
  \caption{MicroXercise System Pipeline: A schematic representation of the exercise analysis pipeline. The process begins with the users performing several repetitions of an exercise, pre-segmented into individual repetitions (light blue). The anchor exercise, considered the ground truth, is colored in orange. The data undergoes a series of processing steps, including primitive (noise) removal, adaptive DTW, and micro-segmentation. In parallel, utilizing \revise{an existing comparative neural network (such as Siamese Neural Network)}, MicroXercise Monitoring produces an attribution map from attribution-based methods. It aids in the video generation process using inverse kinematics and in the text generation. The final video emphasizes key features pinpointed by the attribution map with in-depth, granular feedback on their exercise metrics.}
  \label{fig:sample_run_through}
\end{figure}
\subsection{Overview} 
As shown in Fig. \ref{fig:sample_run_through}, initially, users perform repetitions of an exercise. Each repetition is pre-segmented on the device, resulting in a set of signal exercises in light blue corresponding to individual repetitions. Alongside these, we have the anchor exercise in orange, considered the ground truth. This data then undergoes micro-motion syncing, involving primitive removal, adaptive DTW, and micro-segmentation. Concurrently, MicroXercise Monitoring is performed using a \revise{deep neural network, specifically spatiotemporal SNN,} with attribution-based methods, generating an attribution or an importance heatmap \revise{of the two inputs}. The output image from this process is then used for video generation using inverse kinematics and Levenberg-Marquardt algorithm. Once the video is created, we extract the important features as highlighted by the attribution score and incorporate them into the generated video, segmented into different phases of the exercise. This gives the user detailed, micro-level observational feedback on different exercise metrics.

\subsection{Micro-motion Syncing Analysis}
\label{subsec:micro-motion}
As illustrated in Fig. \ref{fig:sample_run_through}, this section focuses explicitly on the purple area of the diagram, representing the stages of primitive removal, adaptive DTW, and micro-segmentation. These processes are examined in the context of the overarching purpose diagram, moving from top to bottom.
\subsubsection{Primitive Removal}
Primitive removal is a critical process in our system, aimed at eliminating noise in the data to enable accurate calculation of various metrics for a given time series in assigned exercises. This includes removing noise for precise measurement of metrics like the range of motion. Such noise can arise from factors such as sensor inaccuracies during data collection. 
For this purpose, we adopted two techniques: Butterworth low-pass filtering and moving average smoothing.

The Butterworth low-pass filter was chosen for its smooth frequency response, effectively preserving the low-frequency components crucial to our analysis. High-frequency noise can significantly distort measurements. By applying this filter with a threshold frequency of 20 Hz, as suggested by \cite{khusainov_real-time_2013}, we ensure the retention of only those frequencies relevant to our exercise metrics.

Subsequent to the initial noise reduction, we processed the data further using moving average smoothing \cite{roenneberg_chapter_2015}. This method aids in reducing random fluctuations and smoothing out short-term irregularities in the time series data. Sporadic spikes or drops that do not represent actual exercise performance are eliminated by averaging over a specified window.


\begin{figure*}[ht]
    \centering
    \includegraphics[width=\textwidth]{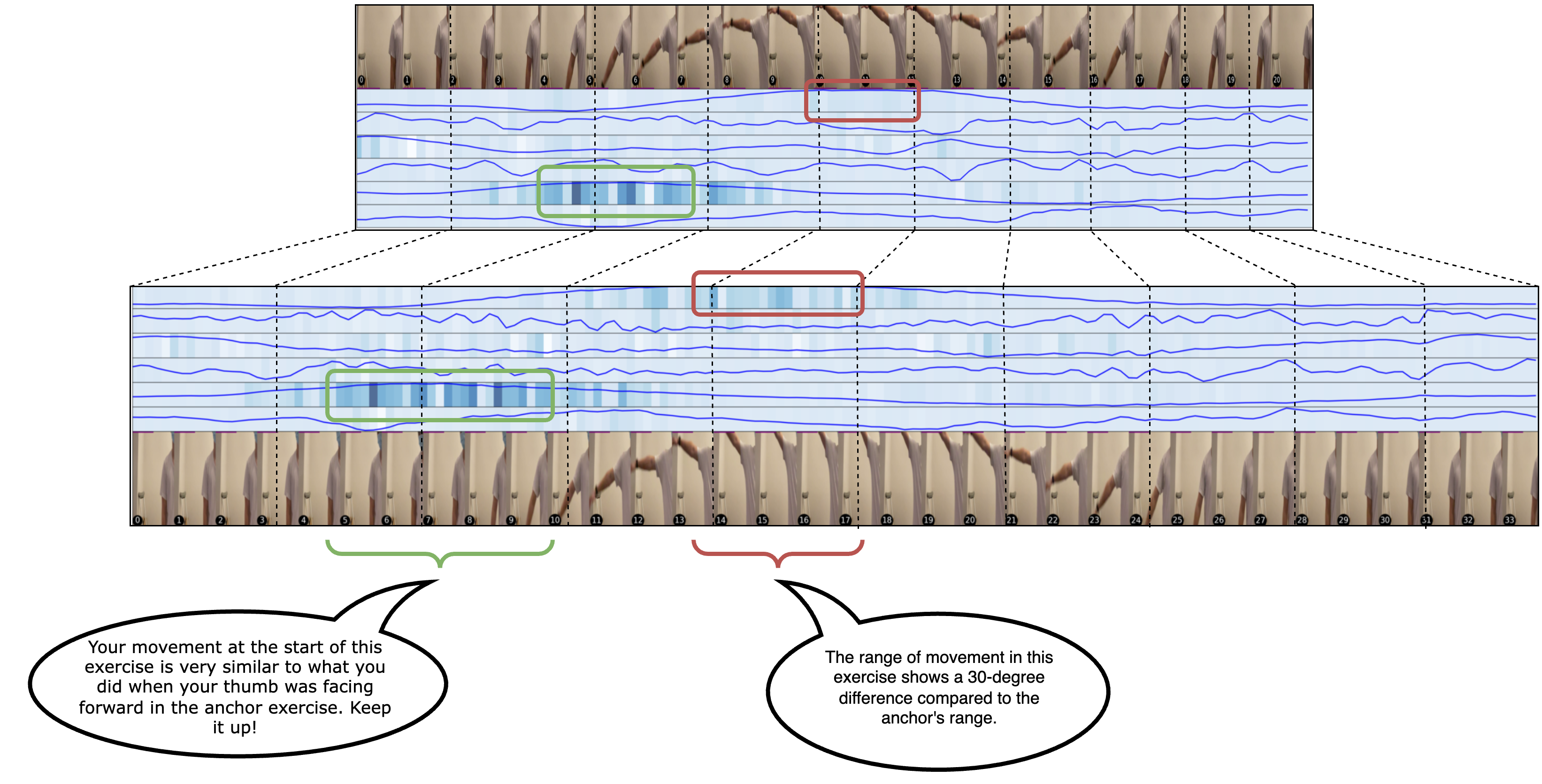}
    \caption{An illustrative diagram to show the comparison of the explainable AI system with Adaptive DTW and segmentation on signal (top) and anchor (bottom) exercises, with video recording, in one repetition. This diagram has seven rows: 3 axial accelerometer, 3 axial gyroscope,  and reference video recording. The signals are also shown in blue, and heatmaps are shown in dark blue.} 
    \label{fig:comparison_xai}
\end{figure*}

\subsubsection{Adaptive DTW}

In this part, we address the challenge of aligning data from different sources to enhance the interpretability and utility of the outputs from our SNN model. Our focus is on using DTW as a tool to align signals, particularly in the context of understanding the network's attributions.

Our analysis primarily utilizes 6-axis input data, consisting of accelerometer and gyroscope measurements from a smartwatch. This decision is influenced by prior research, such as the studies by Burns et al. \cite{burns_shoulder_2018}, and Weiss et al. \cite{weiss_smartwatch-based_2016}, which effectively used these data dimensions for activity recognition and health monitoring. Despite the absence of magnetometer data, these studies demonstrated robust performance, which we aim to emulate and build upon.

To align the signals effectively, we adopted DTW for calculating the distance and path between two time series shown in Fig. \ref{algo:multidim-dtw}, represented as matrices $s$ and $t$. The algorithm computes a distance measure between these matrices, summing the absolute differences between corresponding elements (across the 6-axis data), and is optimal in temporal data.

The significance of employing DTW in our study lies in its ability to bridge the gap between the raw input data and the attributions provided by the SNN. As shown in Figure \ref{fig:comparison_xai}, it provides a visualization of the algorithm we use for this purpose and includes a display of attributions heatmap.  While the neural network operates as a black-box model, offering some insights into its internal workings, DTW provides a tangible means to understand how the input data correlates with the attributions generated. 
\begin{algorithm}
\small
\caption{Multi-dim DTW Distance and Path}
\begin{algorithmic}[1]
\Function{dtw\_dist\_path\_multi}{$s$, $t$}
    \State $n, m \gets \text{length}(s), \text{length}(t)$
    \State $num\_axes \gets \text{number of axes in } s$
    \State $\text{Init } dtw \text{ with } (n+1) \times (m+1) \text{ elems set to } \infty$
    \State $dtw[0][0] \gets 0$
    \For{$i \gets 1 \text{ to } n$}
        \For{$j \gets 1 \text{ to } m$}
            \State $cost \gets \sum_{k=0}^{num\_axes-1} \lvert s[i-1, k] - t[j-1, k] \rvert$
            \State $dtw[i][j] \gets cost + \min(dtw[i-1][j], dtw[i][j-1], dtw[i-1][j-1])$
        \EndFor
    \EndFor
    \State $\text{Init empty list } path$
    \State $i \gets n, j \gets m$
    \While{$i > 0 \text{ or } j > 0$}
        \State $\text{Append }(i, j) \text{ to } path$
        \If{$i == 0$} \State $j \gets j - 1$
        \ElsIf{$j == 0$} \State $i \gets i - 1$
        \Else
            \State $min\_idx \gets \text{argmin}(dtw[i-1][j], dtw[i][j-1], dtw[i-1][j-1])$
            \If{$min\_idx == 0$} \State $i \gets i - 1$
            \ElsIf{$min\_idx == 1$} \State $j \gets j - 1$
            \Else \State $i \gets i - 1, j \gets j - 1$
            \EndIf
        \EndIf
    \EndWhile
    \State \Return $dtw, path$
\EndFunction
\end{algorithmic}
\label{algo:multidim-dtw}
\end{algorithm}

\begin{algorithm}
\small
\caption{Micro-segmentation Algorithm}
\begin{algorithmic}[1]
\Function{micro-segmentation}{$s$, $t$, $n_{\text{seg}}=10$, $N=\text{None}$}
    \State $N \gets \lfloor \frac{\text{length(} t \text{)}}{n_{\text{seg}}} \rfloor$ 
    \State $s, t \gets$ centered moving average of $s$, $t$
    \State $path \gets$ distance and path between $s$ and $t$ using Alg. \ref{algo:multidim-dtw}
    \State $res \gets$ empty list
    \State $min_{dist} \gets \infty$
    \For{$seg_{\text{start}} \gets 0$ \text{ to } $len(t)$ \text{ with step } $N$}
        \State $seg_{\text{end}} \gets \min(seg_{\text{start}} + N, len(t))$
        \State $min_i, min_j \gets -1, -1$
            \For{$i, j \in path$}
                \If{$j-1=seg_{\text{start}}$ and $\lvert s[i-1] - t[j-1] \rvert < min_{dist}$}
                    \State $min_{dist} \gets \lvert s[i-1] - t[j-1] \rvert$
                    \State $min_i, min_j \gets i-1, j-1$
                \EndIf
            \EndFor
        \State Append $(min_i, min_j)$ to $res$
    \EndFor
    \State \Return $res$
\EndFunction
\end{algorithmic}
\label{algo:micro-segmentation}
\end{algorithm}

\subsubsection{Micro-Segmentation}
As described in Algorithm \ref{algo:micro-segmentation}, this function applies a centered moving average to both $s$ and $t$, computes the path using the DTW algorithm in Algorithm \ref{algo:multidim-dtw} between them, and then segments the aligned $t$ sequence into micro-segments of length $N$. For each micro-segment, the function finds the element of $s$ with the minimum distance to the micro-segment using the argmin function, which selects the minimum element from a list of distances computed as the absolute difference between corresponding elements $s$ and $t$ in multi-dimensions. Finally, the function returns \revise{a list res containing pairs of indices. Each pair corresponds to a point in $s$ and a point in $t$ that are closest at the beginning of each segment in $t$.}


\subsection{MicroXercise Monitoring} \label{subsec:xai_monitor}
\subsubsection{Multi-task Siamese Neural Network (SNN)}

\revise{Though we are not limited to using only one type of neural network, to incorporate a complete system, we utilize a comparative model of SNN} to evaluate physical exercise quality. We \revise{adopt and follow} the multi-task \revise{spatiotemporal SNN} structure and implementation in this work \cite{wang_physiq_2023}. The architecture of the SNN is a \revise{combination} of LSTM, CNN, and attention mechanisms with two sub-identical networks. LSTM layers handle the sequential nature of sensor data, while CNN layers extract relevant features. The attention mechanism focuses on significant segments of the data, enhancing the model's interpretative capability. The network employs cosine similarity to measure the closeness of the input exercise to the standard or ``anchor'' exercise.  Additionally, one sub-identical pipeline from SNN is outputting a classification score to have an absolute quality assessment.

For labeling, we rely on annotations from fitness experts. These labels indicate the correctness of exercise execution and serve as a reference for supervised learning. The model's performance is evaluated using Mean Absolute Error (MAE) and R-squared metrics, with Mean Squared Error (MSE) and Cross Entropy as the loss function, focusing on the similarity of the signal to the anchor exercises. This approach ensures that the SNN effectively discerns the quality of physical exercises, providing a reliable tool for fitness assessment.

\subsubsection{Attribution-based Methods}


In our system, we aim to enhance model transparency and comprehensibility by incorporating three distinct attribution techniques, each chosen for its unique strengths in analyzing and filtering data related to micro-motions. These methods are IG, Saliency, and Input X Gradient, and they are employed to assess and refine our model's focus on critical movement features.

The Integrated Gradients (IG) method, outlined by \cite{sundararajan_axiomatic_2017}, excels in providing a detailed analysis of feature importance, crucial for examining micro-motions by quantifying each feature's contribution to model predictions. In contrast, the Saliency method, as described by \cite{simonyan_deep_2014}, offers rapid evaluation, efficiently identifying key input features, thereby streamlining the feature filtering process. Furthermore, the Input X Gradient method, detailed by \cite{shrikumar_not_2016}, is particularly effective in high-dimensional, sparse datasets, focusing on the most influential features to enhance the model’s accuracy in micro-motion analysis.

Collectively, these attribution techniques serve not just to test the model individually but also to refine the input data by emphasizing the most influential features for our micro-motion analysis. This approach ensures that the data fed into our model is of high quality and relevance, enabling more accurate evaluations as we further detail in Sec. \ref{sec:system_eval}.

\subsubsection{Attribution Extraction}

Attribution extraction uses an attribution heatmap  from attribution-based methods, wherein a top $T$ threshold percentage is employed to identify the most important features. Furthermore, we normalize the attribution derived from both signal and anchor exercises collectively. In addition, we leverage the outcomes of \textit{micro-segmentation} to align the top $T$ percent attribution indices. For instance, if the top $T$ percent attribution falls within the third segment, we opt to analyze the third segment rather than the indices.

Additionally, we utilize this result of threshold segmentation to modify the original attribution as we observe that most attribution results are noisy, but based on our method of comparing in the micro-motion, we can signify the signal data while making the anchor data the ground truth for the current comparison. We also further evaluate such an approach with baselines in Sec \ref{sec:system_eval}.

Furthermore, from prior knowledge in the context of the range of motion, the most significant attributions are typically found in the middle of the data sample. This is because the middle region represents the highest degree of motion changes in the supervised learning model, which is logical given that the range of motion (ROM) model is trained to classify various motion ranges. However, this is not applicable if the neural network is trained and evaluated on stability. In this case, we do not assume the location and consider that multiple areas could be on the top $T$ percentage of the attribution. 

\subsection{Generation of Explainable Micro-Motion Feedback} \label{subsec:gen_xai_mmm_res}

\subsubsection{Signal Translation}

In signal translation, we convert the analog value from signal space to physical space to make it more meaningful to users. For example, in a range of motion metric, the usefulness of an interpretation is determined by a degree of motion difference between the signal and anchor exercise. This implies that the data collected from the current exercise should be compared to the data collected from the anchor exercise. 

Consequently, Euler angle estimation \cite{zhou_novel_2021} from accelerometer and gyroscope data is a crucial method to determine the orientation of an object in 3D space. Accelerometers can measure linear acceleration, whereas gyroscopes measure angular velocity. By integrating these measurements, it is possible to determine the orientation of an object in terms of Euler angles (roll, pitch, and yaw).

Accelerometer readings are converted to meters per second squared by multiplying each axis by the standard gravitational acceleration constant (9.81 $m/s^2$). Then trigonometric functions, specifically arcsine and arctangent, derive the pitch and roll angles directly from the normalized accelerometer data. However, the yaw angle cannot be calculated from accelerometer data alone, as additional information is required, such as magnetometer or gyroscope. However, this suffices as it can provide enough insight for exercises involving the difference in range of motion along the primary axis. By integrating the angular velocities over time with pitch and roll angles, the framework acquires the complete set of Euler angles utilized for the end users.

Building on existing research, we propose a measure of stability that takes into account the physical context and characteristics of exercises. This measure, inspired by the work of Yan et al. \cite{yan_normalized_2000}, quantifies hand movement jerk over time, providing a nuanced and physically meaningful measure of stability. We further refine this measure, adopting the Normalized Jerk Score (NJS) proposed by Kitazawa et al. \cite{kitazawa_s_quantitative_1993}, which removes the influence of movement length and duration. The NJS, a unit-free metric, has proven effective in categorizing deviation from a smooth movement \cite{hogan_sensitivity_2009}. The new modified NJS is calculated as follows:

\begin{equation}\label{eq:njs}
    \text{NJS} = -log\left( \Bigl|\frac{\tau^3}{A^2} \sum_{i} (jerk_i)^2 dt \Bigr| \right)
\end{equation}

where $A$ is the peak movement amplitude per axis, $\tau$ is the total duration of the movement, $jerk_i$ is the jerk at $i$ time step, \revise{or represents the second derivative of the position with respect to time}, and $dt$ \revise{is the one over sampling frequency, or time step between consecutive samples.} By normalizing using both the movement's peak amplitude and its total duration, the jerk score is rendered dimensionless. This dimensionlessness is crucial, as it facilitates a direct comparison between movements of diverse characteristics in stability.

\begin{figure}
    \centering
    \subfigure[Overall \label{fig:mode:raw}] {\includegraphics[width=0.32\linewidth]{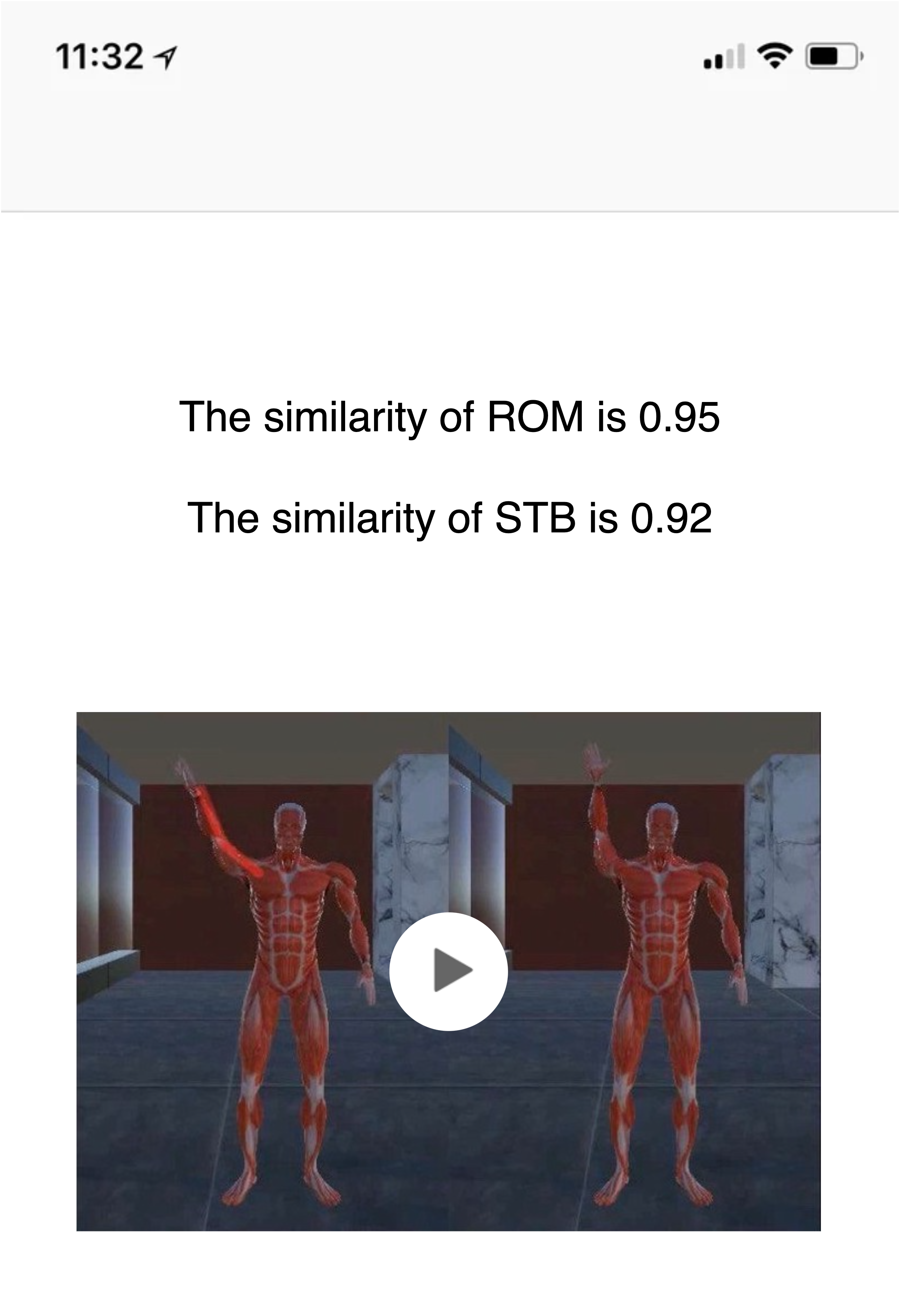}}
    \subfigure[STB   \label{fig:mode:stb}]{\includegraphics[width=0.32\linewidth]{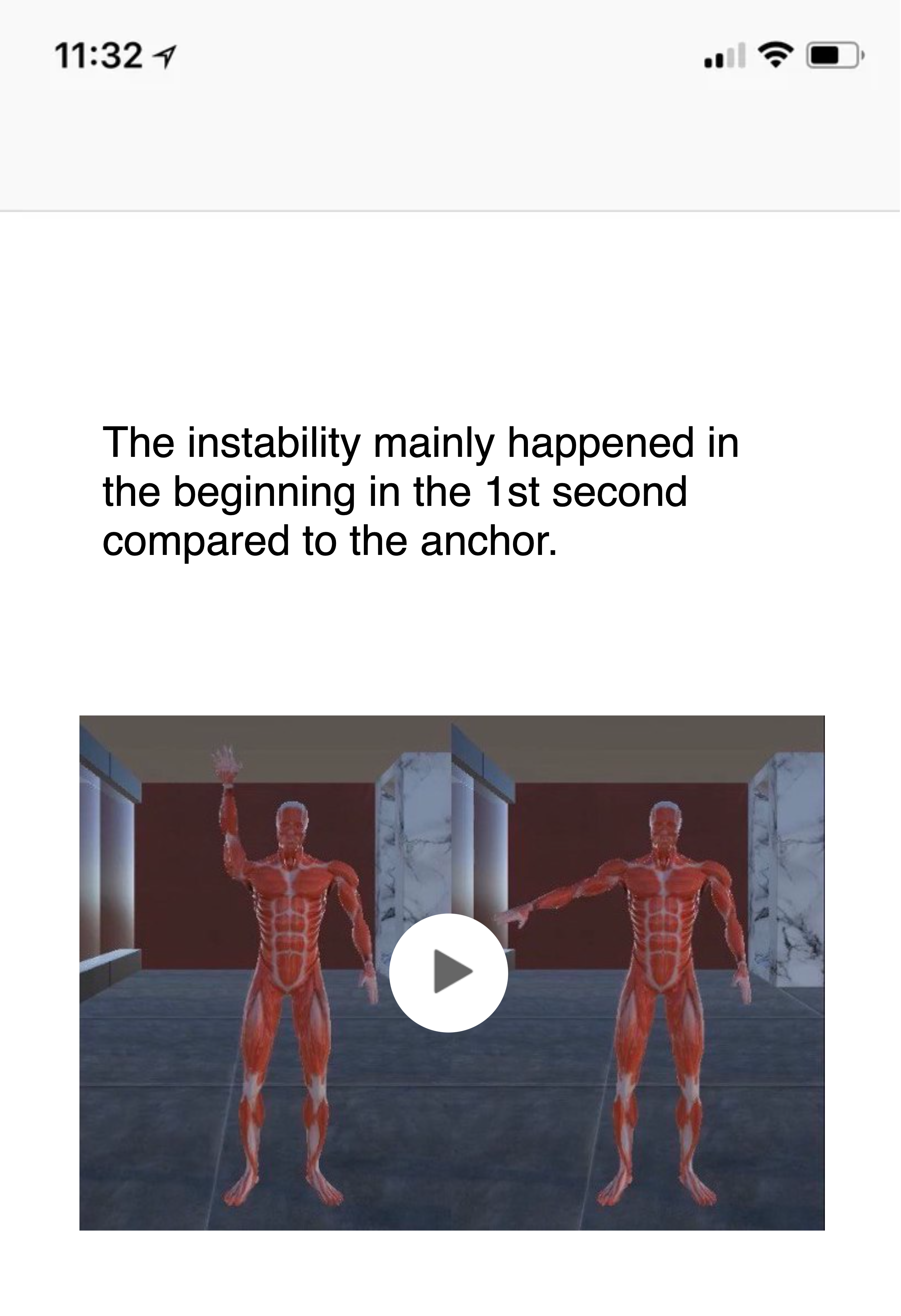}}
    \subfigure[ROM  \label{fig:mode:rom}]{\includegraphics[width=0.32\linewidth]{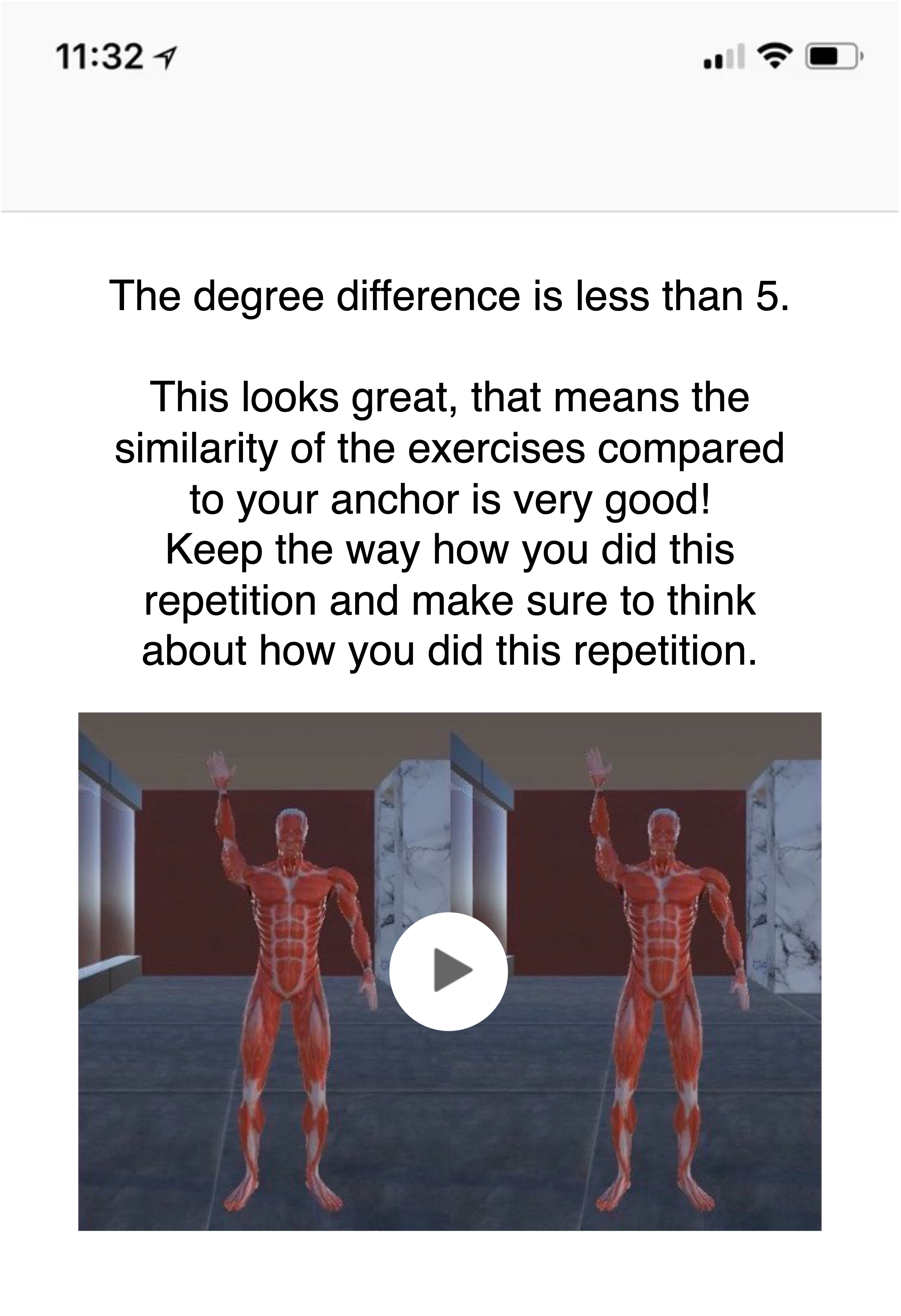}}
    
    \caption{Visual Explainable Results: This result encapsulates critical elements such as the user's similarity score relative to the anchor exercises, temporal fluctuations in stability, and the discrepancy in range of motion at the apex of particular movements. These insights are derived from post-processed attributions generated by our micro-motion analysis.}
    \label{fig:avatar_metrical_modes}
\end{figure}

\subsubsection{Micro Video Generation}
Next, we present the micro video avatar generation system that utilizes IMU data from a smartwatch to reconstruct micro-motion shoulder movements during physical therapy exercises to emphasize the result of micro-motion analysis from attribution maps, as shown in Fig. \ref{fig:avatar_metrical_modes}. The system employs inverse kinematics to solve for the positions of the shoulder and elbow joints in a 3D space. We then use this information to compute the position of the wrist in space relative to a fixed reference frame. To reconstruct the motion of the shoulder and elbow joints in a 3D space, the avatar generation system utilizes a mathematical model of the human arm that includes the shoulder, elbow, and wrist joints. The model assumes that the shoulder joint is a ball-and-socket joint, while the elbow joint is a hinge joint. The angles between the segments of the arm are assumed to be constant, and the lengths of the segments are known.

Using the IMU data and the mathematical model of the arm, the system employs the Levenberg-Marquardt algorithm to solve for the positions of the shoulder and elbow joints. The algorithm minimizes the difference between the actual position and orientation of the end effector, which in this case is the wrist, and the desired position and orientation. The Gauss-Newton algorithm is used to solve for the optimal joint angles when the error between the actual and desired positions is small, while the steepest descent method is used when the error is large. This combination provides a balance between speed and accuracy in solving for optimal joint angles. The following equation can describe the algorithm:

\begin{equation}
\mathbf{J}^T \mathbf{J} \Delta \mathbf{x} + \lambda \mathbf{I} \Delta \mathbf{x} = -\mathbf{J}^T \mathbf{f}(\mathbf{x}),
\end{equation}

where $\mathbf{x}$ is the vector of joint angles, $\mathbf{f}(\mathbf{x})$ is the vector of residuals between the measured and predicted joint positions, $\mathbf{J}$ is the Jacobian matrix of partial derivatives of $\mathbf{f}(\mathbf{x})$ with respect to $\mathbf{x}$, $\Delta \mathbf{x}$ is the update vector for $\mathbf{x}$, $\lambda$ is the damping parameter, and $\mathbf{I}$ is the identity matrix. 

Lastly, as shown in Fig. \ref{fig:avatar_metrical_modes}, our system features three distinct visualization modes: Overall, Stability (STB), and ROM. These modes provide users with a side-by-side comparison of their exercise performance against the original anchor exercises. In this way, users can immediately look into how well they are doing in relation to supervised benchmarks.

\subsubsection{Text Generation}
Building on the visual feedback mechanism, our system incorporates a sophisticated text generation strategy to complement the visual insights. Notably, the same Figure \ref{fig:avatar_metrical_modes} that shows the avatar-based replay also serves as an interface for real-time textual feedback. Our text generation leverages a template-based approach, allowing for concise, modifiable, and quick communication. 

Textual feedback in our system serves as an interactive and intuitive tool designed to guide users constructively. It acts as a valuable, real-time source of advice, providing granular insights into how users perform. For instance, as shown in Fig. \ref{fig:mode:rom}, when a user sees ``The degree difference is less than 5'', it highlights the technical part of precision in their movement, which potentially is useful to the therapists. Positive reinforcement is equally important as plain language, and messages such as ``This looks great. That means the similarity of the exercises compared to your anchor is very good!'' encourage and motivate users by acknowledging their progress. Furthermore, feedback like ``No need to modify the way you do it'' provides affirmation and potential improvement, assuring users that their current method is effective. Overall, our system's textual feedback is designed to support, guide, and foster confidence in users, enabling them to make the most of their exercises and routines.








\section{Evaluation}\label{sec:system_eval}
Assessing attribution methods is crucial to verify their effectiveness and applicability. Our MicroXercise system, an amalgamation of signal processing in micro-motion analysis and an attribution-based deep learning model, seeks to advance the saliency heatmap by presenting refined and nuanced attributions. We present a  quantitative evaluation that provides an empirical foundation for our approach with defined metrics that focus on the objectives of achieving both \textit{fidelity} and \textit{interpretability} in the realm of attribution-based methods \cite{lopes_xai_2022, rojat_explainable_2021, linardatos_explainable_2020, joshi_review_2021}.

\begin{figure*}[htbp] 
  \centering
  \includegraphics[width=1\textwidth]{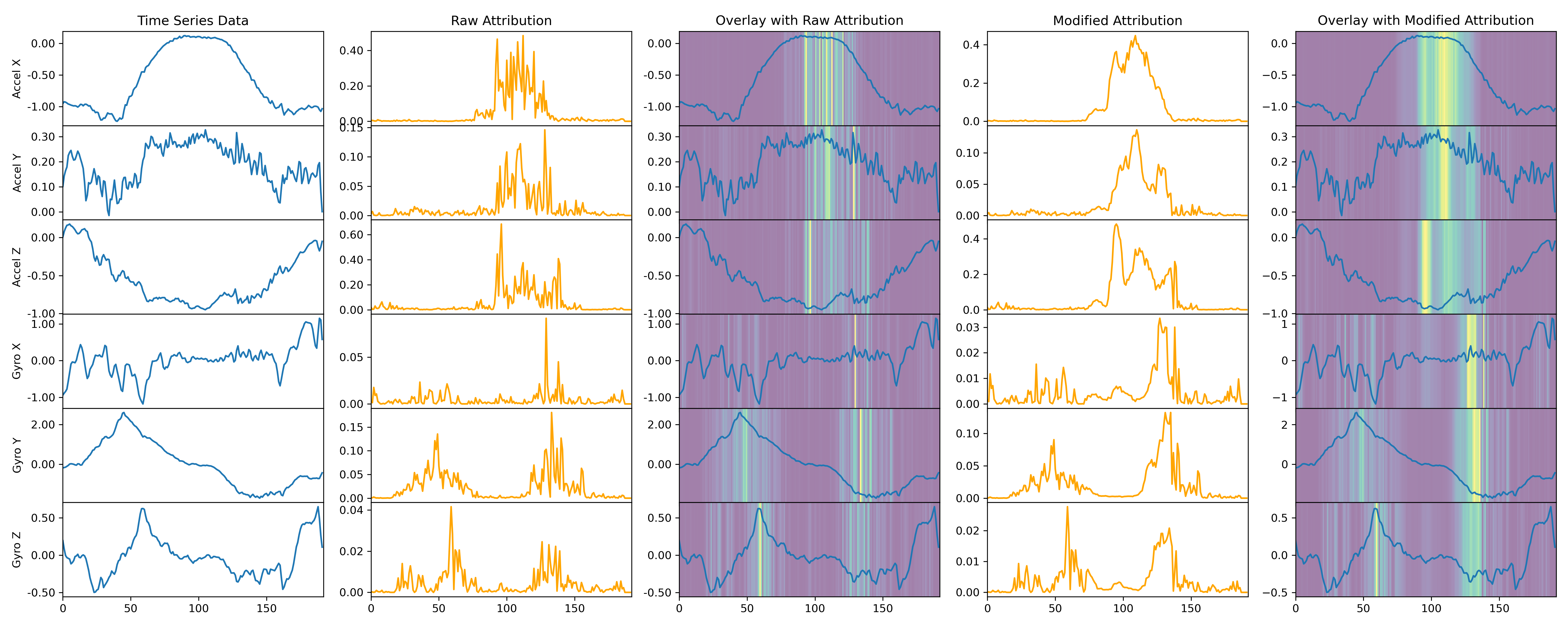} 
  \caption{Signal comparison with raw attribution versus modified attribution in the attribution produced by IG. As shown in the figure, column 1 shows its original signal exercise. The raw attribution is very noisy and inconsistent in column 2 and 3, but the modified attribution produces more consistent results as shown in column 4 and 5.} 
  \label{fig:signal_comparison}
\end{figure*}
\subsection{Dataset and Evaluation Setup}
The dataset used in this evaluation is adapted from \cite{wang_physiq_2023}, \revise{which is consists of multiple shoulder physical therapy exercises}. 
\revise{The dataset is collected using consumer-grade iOS Apple Watches for three exercises. These exercises are chosen because they demonstrate repetitive nature, have clear start and end points, can potentially improve the body, and engage various muscle groups. With the supervision of exercise expert, the participants perform a number of sets of exercises with various range of motion and repetitions with variations of stability. The dataset includes data from 17 male and 14 female participants, aged between 18 and 44, including 3 participants with self-reported previous shoulder injuries.} 

\revise{We train the comparative deep neural networks to compare and interpret feedback, particularly using two exercises of shoulder abduction and forward flexion from the dataset, which contains 1,550 segmented one-repetition exercises.} \revise{The shoulder abduction exercise is collected with 5 range of motions. The forward flexion is collected similarly with same range of motions.}

For the \revise{model}, every possible \revise{pair} of inputs was methodically generated. These \revise{pairs} are associated with a discrete and continuous score of quality assessment, sourced from range-of-motion and stability labels, which established the target for the SNN. The data \revise{are} split into 70\% for training, 10\% for validation, and 20\% for testing in both range-of-motion and stability \revise{metrics}.

We refine the attributions map, produced by attribution methods, using our \textit{Micro-Motion Syncing} methods to produce a modified attribution map. As shown in Fig. \ref{fig:signal_comparison}, by integrating this into the attribution as a layer of prior knowledge, we essentially amplified the significance within crucial segments using a signal smoothing factor. 

\subsection{Evaluation Metrics}
\subsubsection{Fidelity}
Within the realm of fidelity, \textbf{monotonicity} plays a essential role. The fundamental premise is to ensure that as the importance of a feature amplifies, so does its attribution, and vice-versa. This behavior is quantified by computing correlation coefficients, Spearman's rank correlation coefficient, between the feature importance and their corresponding attributions. A strong positive correlation implies a desirable monotonic behavior, attaching to the soundness of our explanations. Adopted by \cite{nguyen_quantitative_2020}, the monotonicity metric is defined as 

\begin{equation}
\rho_S(a, e) = 1 - \frac{ \sum (rank(a_i) - rank(e_i))^2}{n(n^2 - 1)}
\end{equation}

where \( \rho_S \) represents the Spearman’s correlation coefficient, \( a \) is a vector containing the absolute values of the attributions for each feature, 
denoted as \( a = (\ldots, |a_i|, \ldots) \), and \( e \) is a vector containing the expected losses when considering each feature with other features held constant. The $rank$ refers to the numerical ordering of each element within the flattened arrays of attributions and expected losses by their size.


\subsubsection{Interpretability}
In transitioning to interpretability, 
it's critical to ensure that our method doesn't overly focus on specific features while also confirming that the explanation isn't unduly complex. Therefore, in interpretability, we want to focus on \textbf{feature mutual information} and \textbf{continuity}.

Feature mutual information between the original feature sets and their corresponding explanations serves as an apt metric for this purpose. An optimal mutual information score indicates a harmonious blend of broadness and simplicity in our explanations \cite{nguyen_quantitative_2020}.
\begin{equation}
I(x, \alpha) = \sum_{x \in X, a \in \alpha} p(x,a) \log \left( \frac{p(x,a)}{p(x)p(a)} \right)
\end{equation}

Where \( p(x,a) \) is the joint probability distribution of \( x \) and \( \alpha \), \( p(x) \) and \( p(a) \) are the marginal probability distributions of them, respectively. A high value of \( I(x, \alpha) \) indicates that the extracted features (in this case, attributions) retain a significant amount of information from the original input, thus ensuring fidelity in our system. Lastly, we estimated mutual information using a histogram-based approach with 200 bins, because the goal of assessing feature-attribution fidelity is informative alignment and efficiency in multivariate time series data. 

The continuity of an explanation is an another essential method for understanding its usefulness and reliability. For a prediction function \( f(x) \), which we assume to be continuous, the continuity of its explanation is defined by ensuring that similar data points result in closely resembling explanations. Mathematically, the continuity of the explanation, \( \alpha \), or attribution, is quantified by evaluating the most substantial variation in the explanations over the input domain \cite{montavon_methods_2018}. This is represented as:

\begin{equation}
\text{Continuity}(x, x', \alpha, \alpha') = \max_{x \neq x'} \frac{\|\alpha - \alpha' \|_1}{\|x - x'\|_2}
\end{equation}

Here, the numerator measures the difference between the explanations for two data points \( x \) and \( x' \), while the denominator captures the difference between the data points themselves. Additionally, \(x'\) is  the given adjacent values around \( x\) and similarly with \( \alpha'\) around  \( \alpha\). We set 5 adjacent neighbors on each side. A low value of this metric indicates that the explanation is continuous, implying that minor changes in the input lead to proportionate alterations in the explanation, providing clarity and consistency in the interpretation. 



\begin{table}[t]
  \centering
  \caption{Comparison of Metrics for Different Methods and Exercises (Monotonicity (Mono) higher is better, FMI higher is better, and Continuity (Cont) lower is better, FMI has a scaling factor of 10000. XG is InputXGradient, SA is Saliency, and IG is Integrated Gradient. Our method is shown on the second row for each metric for each method modified as MicroXercise, compared with its baseline.)}
  \begin{adjustbox}{max width=\columnwidth}
    \begin{tabular}{ccc||c||c||c}
    \toprule
    & & \multicolumn{2}{c}{Shoulder Abduction} & \multicolumn{2}{c}{Forward Flexion} \\
    \cmidrule(lr){3-4} \cmidrule(lr){5-6}
    Method & Metric & Stability & Range of Motion & Stability & Range of Motion \\
    \midrule
    \multirow{6}{*}{XG} & Mono & \textbf{0.561} & \textbf{0.571} & \textbf{0.624} & \textbf{0.630} \\
                        & Mono (MicroXercise) & 0.505 & 0.465 & 0.328 & 0.572 \\
                        \cmidrule{2-6}
                        & FMI & 87.915 & 94.137 & \textbf{182.091} & 122.859 \\
                        & FMI (MicroXercise) & \textbf{127.170} & \textbf{137.868} & 104.333 & \textbf{172.201} \\
                        \cmidrule{2-6}
                        & Cont & 13.833 & 13.270 & 13.679 & 14.992 \\
                        & Cont (MicroXercise) & \textbf{8.340} & \textbf{5.723} & \textbf{9.992} & \textbf{6.597} \\
    \midrule
    \multirow{6}{*}{SA} & Mono & \textbf{0.679} & \textbf{0.685} & \textbf{0.727} & \textbf{0.711} \\
                        & Mono (MicroXercise) & 0.582 & 0.527 & 0.274 & 0.591 \\
                        \cmidrule{2-6}
                        & FMI & 44.719 & 48.923 & \textbf{69.495} & 51.823 \\
                        & FMI (MicroXercise) & \textbf{58.893} & \textbf{62.512} & 57.239 & \textbf{68.258} \\
                        \cmidrule{2-6}
                        & Cont & 25.592 & 25.916 & 26.886 & 30.106 \\
                        & Cont (MicroXercise) & \textbf{15.593} & \textbf{11.865} & \textbf{16.696} & \textbf{13.500} \\
    \midrule
    \multirow{6}{*}{IG} & Mono & 0.299 & \textbf{0.415} & \textbf{0.519} & \textbf{0.541} \\
                        & Mono (MicroXercise) & \textbf{0.300} & 0.392 & 0.309 & 0.514 \\
                        \cmidrule{2-6}
                        & FMI & 182.669 & 650.714 & \textbf{458.273} & 198.494 \\
                        & FMI (MicroXercise) & \textbf{268.382} & \textbf{924.050} & 319.074 & \textbf{282.383} \\
                        \cmidrule{2-6}
                        & Cont & 10.988 & 7.809 & 11.618 & 13.049 \\
                        & Cont (MicroXercise) & \textbf{6.715} & \textbf{3.455} & \textbf{8.260} & \textbf{5.486} \\
    \bottomrule
    \end{tabular}%
    \end{adjustbox}
\label{tab:comparison}%
\end{table}%

\subsection{Results}
Given the extensive size of our dataset, evaluating the performance of attribution methods on each sample would be computationally expensive and time-consuming. This complexity is particularly exacerbated in our case, where the input data consists of continuous signals. Perturbation-based attribution methods, which require manipulations at each data point, significantly increase computational costs.

To mitigate this, we adopted a random sampling strategy of selecting 100 random sample pairs for each subject under study. This sub-sampling approach allowed us to perform a comprehensive yet manageable evaluation.

\subsubsection{Range-of-Motion}
Table \ref{tab:comparison} presents a comprehensive evaluation of stability and range of motion metrics for both shoulder abduction and forward flexion exercises. It compares our micro-motion-enhanced method with three baseline attribution methods: inputXgradient (XG), Integrated Gradient (IG), and Saliency (SA). Noted, these methods require a `baseline' to perturbate against the output; our choice of baseline value is randomly generated on a normal distribution. While the table reveals somewhat consistent outcomes across the methods, a closer look offers nuanced insights in the range of motion. Specifically, our method shows a reduction in monotonicity, which might initially suggest decreased performance in the range of motion. However, this reduction in `flow' between the time series data and the modified attribution is not necessarily a drawback. 

\begin{figure*}[ht]
\centering
\subfigure[Monotonicity \label{fig:bar:mono}] {\includegraphics[width=0.3\textwidth]{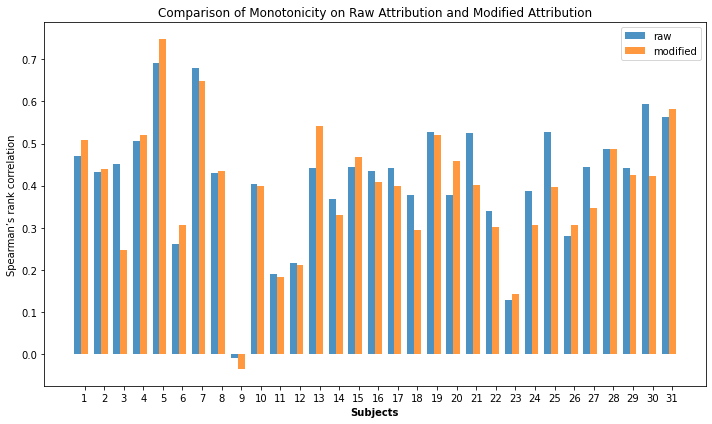}}
    \subfigure[Feature Mutual Information \label{fig:bar:fmi}]{\includegraphics[width=0.3\textwidth]{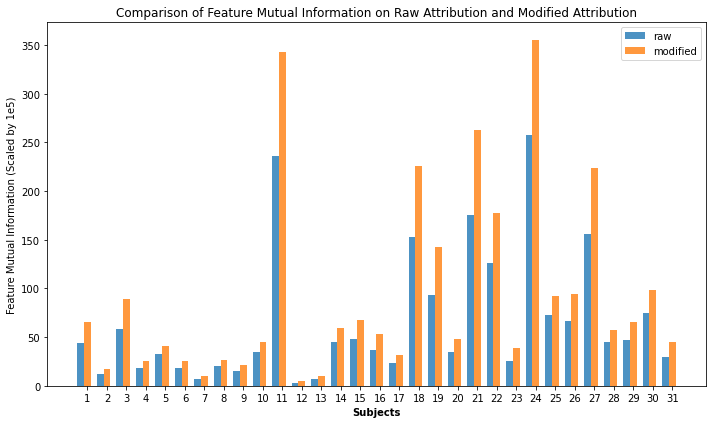}}
    \subfigure[Continuity \label{fig:bar:cont}]{\includegraphics[width=0.3\textwidth]{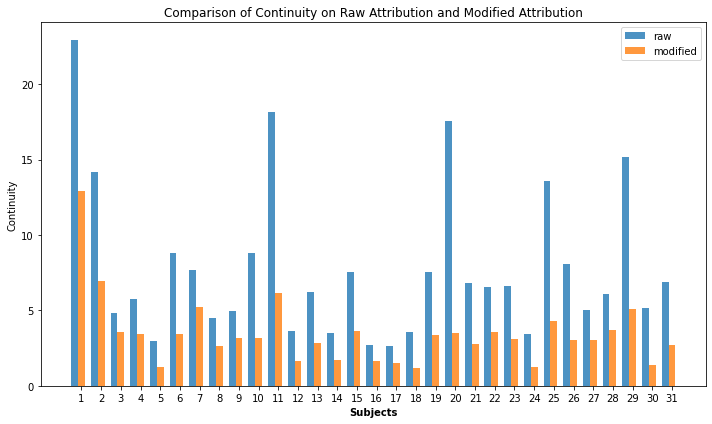}}
    \caption{Comparison of raw and modified attribution on each subject for shoulder abduction on IG. As shown in the figure, our method performs significantly better in Feature Mutual Information and Continuity than the baseline.}
    \label{fig:bar_graph1}
\end{figure*}

As shown in Fig \ref{fig:bar:mono}, the monotonicity merely decreases from the baseline, which is a good sign as we are attempting to modify or smooth the attribution to be more representative to the users. One possible reason to the decrease is we perform the primitive removal to smooth the data followed by micro-motion analysis and segmentation. This analysis is based on the smoothed data which could make the attribution less correlated with the raw input data. Again, our goal is to present the users with more expressive explanation. Additionally, as evidenced by the FMI and Continuity (Cont) metrics, our approach actually outperforms the baselines consistently. This could be that the modified attributions in our method provide richer information, as indicated by higher FMI values, and greater continuity, making the attributions less noisy and more consistent, as corroborated by Figure \ref{fig:signal_comparison}.

\subsubsection{Stability}
The micro-motion approach has a notable impact on the stability metric of the attribution. On one hand, our method successfully attenuates the high noise levels commonly observed in raw attributions, as indicated by improved stability scores in the table for both shoulder abduction and forward flexion exercises. However, it's important to consider that enhanced stability may not always equate to superior attribution quality. In some instances, the `instability' reflected in the baseline attribution may be an intentional and informative characteristic, representing the model's sensitivity to particular features in the input. Thus, while our method scores higher in terms of stability, it could be argued that this might lead to the removal of certain informative inconsistencies originally present in the raw attributions.

\begin{figure*}[ht]
    \centering
    \subfigure[Monotonicity \label{fig:line:mono}] {\includegraphics[width=0.3\textwidth]{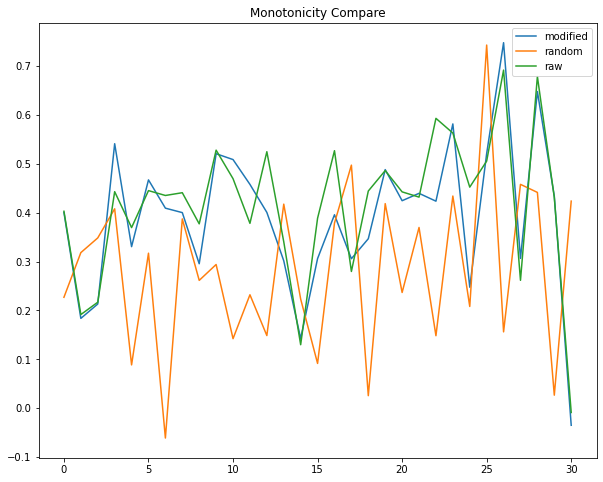}}
    \subfigure[Feature Mutual Information \label{fig:line:fmi}]{\includegraphics[width=0.3\textwidth]{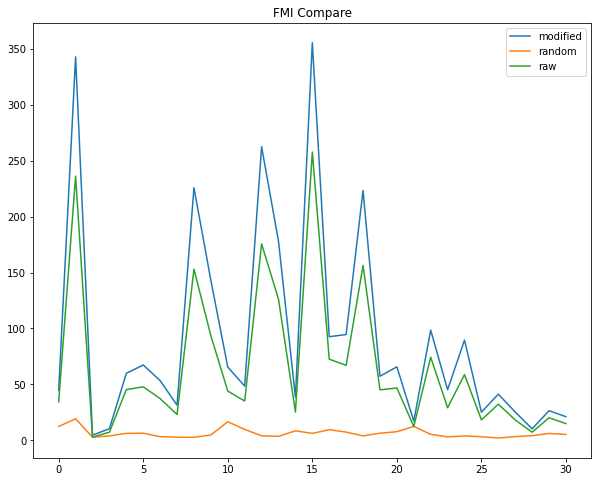}}
    \subfigure[Continuity \label{fig:line:cont}]{\includegraphics[width=0.3\textwidth]{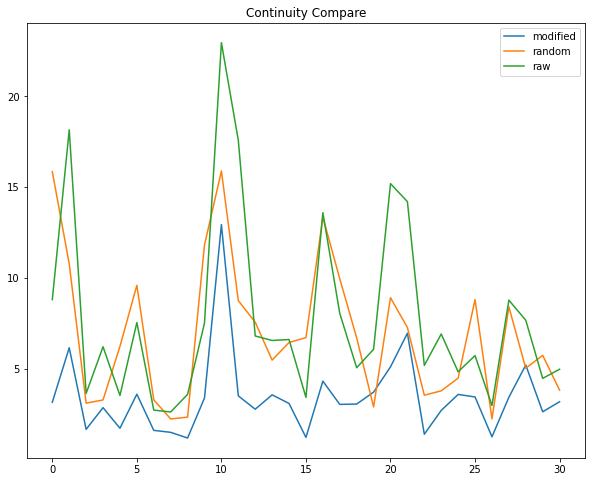}}
    \caption{Comparison of raw (green), modified (blue), and randomized (orange) attribution on each subject for shoulder abduction on Integrated Gradient. As shown in the line figure, our method performs better in Feature Mutual Information (higher is better) and Continuity (lower is better) than the baseline. FMI values are shown factored down by \(1 \times 10^5\) }
    \label{fig:line_graph1}
\end{figure*}

\subsubsection{Randomization}
To validate that the improvements conferred by our micro-motion segmentation and modification are methodologically sound and not attributable to random fluctuations, we devise a comparative experiment. This is critical to demonstrate that our modifications represent a genuine enhancement deriving from principles of signal processing, rather than mere fortuitous events. In this comparative setup, we focus  specifically on the exercise of shoulder abduction and employed the IG method for attribution.

We select 100 pairs of samples for each subject, maintaining similar experimental setup to align with previous evaluations. In addition to applying our micro-motion analysis and segmentation, we also generate a control group by selecting random segments of the same length—25 timestamps, equivalent to half a second—which were not part of the segments our system initially identified as critical. We apply identical modifications to these randomly selected segments on raw attribution, effectively serving as a baseline for performance comparison.

Our results, illustrated in Figure \ref{fig:line_graph1}, suggest a convincing story. In terms of feature mutual information and continuity, our method, represented by the blue curve, distinctly outperforms both the baseline and the randomly modified segments. This validates that the information capture and continuity improvements are not artifacts but are attributable to our system. It's important to note that while there is a similar performance with some declines at times in the monotonicity metric when using our method (and clearly better than baseline), this doesn't impact the overall quality of the attribution as shown earlier, thereby affirming the robustness of our approach.

\revise{Interestingly, feature mutual information shows a reasonable trend that by randomly selecting segments in samples, it shows a significantly lower feature mutual information. But randomly selected segments in monotonicity and continuity still have some high level of numeric results, meaning possibly that monotonicity and continuity emphasize the overall quality of attribution as feature mutual information focuses more on the local attribution. These results further demonstrate that the original authors' claims on feature mutual information capture the property of simplicity and broadness in data with respect to its generated attribution.}

\section{Related Work}\label{sec:related_work}


\subsection{Technologies in Remote Physical Therapy Exercise}


The landscape of remote health interventions is diverse, with several promising technologies emerging in recent years. A systematic review by Grona et al.\cite{grona_use_2018} examined 17 full-text articles applying real-time physical therapy interventions through secure videoconferencing. The results showed general patient satisfaction but highlighted the need for more rigorous study designs. Similarly, Pietrzak et al.\cite{pietrzak_self-management_2013} revealed the effectiveness of internet-based technologies in providing community-based self-management and rehabilitation interventions for osteoarthritis patients. On the other hand, mHealth apps for remote health, such as the one proposed by Burns et al.\cite{burns_shoulder_2018}, focus on classifying exercises without considering the quality of the exercise execution.  Mork et al.\cite{mork_decision_2018} outlined a protocol for designing and implementing a decision support system called selfBACK, which was aimed at promoting self-management of nonspecific low back pain among patients, with their system of case-based reasoning technology. Smittenaar et al. \cite{smittenaar_translating_2017} investigated the effects of the Hinge Health 12-week digital care program on chronic knee pain, function, surgery interest, and satisfaction. The \revise{care program} included sensor-guided physical exercises, weekly education, activity tracking, and psycho-social support, such as personal coaching and cognitive behavioral therapy.


While the advancements in digital health interventions are noteworthy, a significant challenge persists in the form of transparency and explanation of the provided recommendations. For instance, the selfBACK system by Mork et al.\cite{mork_decision_2018} doesn't readily make the reasoning behind its recommendations apparent to users. Similarly, while the Hinge Health DCP, as discussed by Smittenaar et al.\cite{smittenaar_translating_2017}, is effective in improving pain and function and in decreasing surgery interest, it fails to offer XAI feedback. Further, reviews of existing apps underscore the deficiencies in current solutions. For instance, research by Dantas et al. and Agnew et al.\cite{dantas_mobile_2020, agnew_investigation_2022} revealed a dearth of functional, user-centered tools for Systemic Lupus Erythematosus patients, with most apps offering only partial solutions. A review by Carvalho et al.\cite{carvalho_mobile_2022} found that mHealth technologies for managing spine disorders in Brazilian online app stores exhibited acceptable to inadequate quality.  

\subsection{Attribution-based Explainable AI}

In the domain of XAI, attribution-based methods have gained significant attention for their ability to interpret complex models by assigning importance to input features. Notable techniques include Grad-CAM \cite{selvaraju_grad-cam_2020}, which utilizes gradient-based localization to highlight significant regions in input images. Its primary advantage lies in producing high-resolution and class-discriminative visualization, making it suitable for tasks where spatial localization is crucial. Integrated Gradient \cite{sundararajan_axiomatic_2017} offers a more axiomatic approach, providing a path integral over the model's gradients to delineate feature importance. It adheres to foundational axioms like sensitivity, implementation invariance, and completeness, which makes it widely applicable beyond image-based models to even structured data. DeepLIFT (Deep Learning Important FeaTures) \cite{shrikumar_learning_2019} contrasts the activation of each neuron to a `reference activation' to compute the `contribution' score of each feature. Input X Gradient \cite{shrikumar_not_2016} explores the interaction between input and gradient during backpropagation, identifying features that significantly contribute to the output. It can be a simpler yet insightful method, especially when computational resources are limited.

\subsection{Feedback and Visualization Mechanisms}
The application of wearable devices and XAI in healthcare has demonstrated noteworthy advancements across diverse medical domains. Frade et al. employed wearable devices and machine learning to estimate cardiovascular fitness, offering a modern substitute for traditional cardiopulmonary exercise testing \cite{frade_toward_2023}. Their use of Shapley values for attribute importance brings a layer of interpretability to their models \cite{lundberg_unified_2017}. Likewise, Arrotta et al. presented DeXAR, an innovative technique for recognizing activities of daily living in smart-home settings \cite{arrotta_dexar_2022}. They employed attribution methods such as Grad-CAM \cite{selvaraju_grad-cam_2020} based on CNNs to translate chronological activities into a spatial representation, demonstrating the efficacy of white-box XAI methods.

Building on the necessity for transparent decision-making in healthcare, Biswas et al. argued for explainability in AI systems when dealing with Autism Spectrum Disorder datasets \cite{biswas_xai_2021}. Yang et al. furthered the XAI discourse by proposing a multi-modal and multi-center data fusion approach for weakly supervised learning applications in healthcare \cite{yang_unbox_2022}. Their work targets the 'black-box' nature of deep learning algorithms, advocating for clearer understanding of AI-driven decisions.

Slijepcevic and others conducted a thorough evaluation of attribution-based methods in clinical gait analysis, aiming to understand their behavior across various deep learning models \cite{slijepcevic_explaining_2022}. 
Their research sheds light on the value of XAI methods from a clinician's perspective.

While these advancements signify promising progress, a common limitation is their focus on global or macro-level explanations. Unlike these approaches, our MicroXercise system leverages IMUs in the realm of remote physical therapy. By targeting micro-motions, we aim to provide highly detailed and personalized feedback, thereby addressing the gaps in existing XAI methodologies and significantly enhancing the user's therapeutic experience.

\section{Conclusion and Future Work}\label{sec:summary}
\revise{In this work, we introduced MicroXercise, a mobile application leveraging attribution-based methods in synergy with our multi-dimensional DTW to assist users with their PT exercise routines at a granular, micro-motion level. This novel integration shows a step forward in enhancing shoulder exercise effectiveness and feedback for users.}

\revise{Though our results show promise for supporting home-based PT, we also acknowledge that there are limitations of home-based PT, such as environmental constraints, access to professional feedback, human interventions, and system scalability, that must be addressed prior to deploying the application for widespread use. To address these limitations, we plan to extend the work in the following directions. First, in video feedback, we will investigate more accurate and sophisticated algorithms how to visualize users' exercises, such as real-world uncertainties. Second, we plan to conduct a series of user studies to solicit feedback from physical therapists and patients to improve the usability and design of the application, as this will have substantial impact on user adoption, particularly for a health-related technology. Moreover, we intend to broaden our scope of work to include other types of exercises or sports.
Furthermore, ensuring scalability and deployment requires a suite of wearable devices, robust server infrastructure for algorithm processing, educational materials for app utilization, and additional support for various PT exercises.
We will also examine the integration of other sensors or a full-body motion tracking system to provide a more thorough exercise feedback mechanism, as well as ensure that the sensitive health data collected by the system is secured based on HIPAA standards. Finally, we will ensure that ethical considerations related to AI-based recommendations in healthcare be considered prior to widespread dissemination of this research to the public.}

\bibliographystyle{IEEEtran}
\bibliography{nourl_references}

\end{document}